\documentclass[sigconf]{acmart}
\makeatletter                   
\def\mdseries@tt{m}             
\makeatother                    
\usepackage[plain]{fancyref}
\usepackage[draft=true]{minted} 
\usepackage{color}
\usepackage{hyperref}           
\hypersetup{
    colorlinks=true,
    linkcolor=blue,
    filecolor=red,      
    urlcolor=magenta,
    breaklinks=true,            
}
\usepackage{breakurl}           
\usepackage{url}

\usepackage{color}
\usepackage{comment}
\usepackage{url}
\usepackage[utf8]{inputenc}
\usepackage{enumitem} 
\usepackage{float}
\usepackage{balance}
\usepackage{graphicx}
\usepackage{booktabs} 
\usepackage{multirow} 
\usepackage{framed}   
\usepackage{amsmath,bm} 
\usepackage{pgfplots} 
\usepackage{multicol}
\usepackage{subfig} 
\usepackage{algorithm}
\usepackage[noend]{algpseudocode}
\usepackage{makecell} 
\usepackage{pgfplots, filecontents} 
\usepackage{caption}
\usepgfplotslibrary{external,colormaps,groupplots,statistics}
\pgfplotsset{compat=1.8}
\usepgfplotslibrary{statistics} 
\usepackage{soul} 
\usepackage{xspace} 

\newcommand{\ie}{\emph{i.e.,}\xspace}
\newcommand{\eg}{\emph{e.g.,}\xspace}
\newcommand{\etc}{etc.\xspace}
\newcommand{\etal}{\emph{et~al.}\xspace} 
\newcommand{\descStep}[2]{\noindent \textbf{#1: } #2}

\newcommand{\smallTitle}[1]{\vspace{0mm} \noindent \textbf{#1 }} 
\newcommand{\ToolURL}{\url{https://tacticvolatility.github.io/}} 

\newcolumntype{C}[1]{>{\centering\arraybackslash}p{#1}}
\newcommand{\TotalTacticOneServerOneEvals}{52,106~} 

\newcommand{\dan}[1]{\textcolor{blue}{{\it [Dan: #1]}}}
\newcommand{\ahmed}[1]{\textcolor{cyan}{{\it [Ahmed: #1]}}}

\setcopyright{acmcopyright}
\copyrightyear{2022}
\acmYear{2022}
\acmDOI{10.1145/1122445.1122456}

\copyrightyear{2022}
\acmYear{2022}
\setcopyright{acmcopyright}\acmConference[GECCO '22]{Genetic and Evolutionary Computation Conference}{July 9--13, 2022}{Boston, MA, USA} \acmBooktitle{Genetic and Evolutionary Computation Conference (GECCO '22), July 9--13, 2022, Boston, MA, USA}
\acmPrice{15.00}
\acmDOI{10.1145/3512290.3528745}
\acmISBN{978-1-4503-9237-2/22/07}

\begin{document}

\title[Addressing Tactic Volatility with eRNNs and Uncertainty Reduction Tactics]{Addressing Tactic Volatility in Self-Adaptive Systems Using Evolved Recurrent Neural Networks and Uncertainty Reduction Tactics}


\author{Aizaz Ul Haq}
\affiliation{%
  \institution{Rochester Institute of Technology}
  \city{Rochester, NY}
  \country{USA}}
\email{au7149@rit.edu}

\author{Niranjana Deshpande}
\affiliation{%
  \institution{Rochester Institute of Technology}
  \city{Rochester, NY}
  \country{USA}}
\email{nd7896@rit.edu}

\author{AbdElRahman ElSaid}
\affiliation{%
  \institution{University of Puerto Rico}
  \city{Mayagüez, PR}
  \country{USA}}
\email{abdelrahman.elsaid@upr.edu}

\author{Travis Desell}
\affiliation{%
  \institution{Rochester Institute of Technology}
  \city{Rochester, NY}
  \country{USA}}
\email{tjdvse@rit.edu}
\author{Daniel E. Krutz}
\affiliation{%
  \institution{Rochester Institute of Technology}
  \city{Rochester, NY}
  \country{USA}}
\email{dxkvse@rit.edu}

\renewcommand{\shortauthors}{Haq and Deshpande, \etal}

\begin{abstract}
  Self-adaptive systems frequently use \emph{tactics} to perform adaptations. Tactic examples include the implementation of additional security measures when an intrusion is detected, or activating a cooling mechanism when temperature thresholds are surpassed. \emph{Tactic volatility} occurs in real-world systems and is defined as variable behavior in the attributes of a tactic, such as its latency or cost. A system's inability to effectively account for tactic volatility adversely impacts its efficiency and resiliency against the dynamics of real-world environments. To enable systems' efficiency against tactic volatility, we propose a \emph{Tactic Volatility Aware} (TVA-E) process utilizing \emph{evolved Recurrent Neural Networks} (eRNN) to provide accurate tactic predictions. TVA-E is also the first known process to take advantage of \emph{uncertainty reduction tactics} to provide additional information to the decision-making process and reduce uncertainty. TVA-E easily integrates into popular adaptation processes enabling it to immediately benefit a large number of existing self-adaptive systems. Simulations using \TotalTacticOneServerOneEvals tactic records demonstrate that: I) eRNN is an effective prediction mechanism, II) TVA-E represents an improvement over existing state-of-the-art processes in accounting for tactic volatility, and III) Uncertainty reduction tactics are beneficial in accounting for tactic volatility. The developed dataset and tool can be found at \ToolURL

\end{abstract}




\maketitle

\newcommand{\RQA}{Is eRNN effective for predicting tactic volatility?} 
\newcommand{\RQB}{Does eRNN effectively transfer the tactic volatility predictions to effective decisions compared with other baselines?} 
\newcommand{\RQC}{Are Uncertainty Reduction Tactics effective in helping to predict tactic volatility?}

\section{Introduction}


Self-adaptive systems utilize \emph{adaptation tactics} to respond to events and accomplish system goals~\cite{moreno2017adaptation,angelopoulos2016model}. Example tactics include a web farm provisioning an additional virtual machine (VM) when the workload reaches a specific threshold, or reducing non-essential functionality on an autonomous Unmanned Aerial Vehicle (UAV) when battery levels are low. Many tactics will have attributes in the form of latency and cost. Tactic latency is the amount of time from when a tactic is invoked until its effect on the system is realized~\cite{Moreno:2018:FED:3208359.3149180,Moreno:2017:CMP:3105503.3105511}. Tactic cost is the resources necessary to complete the execution of the tactic, where `cost' could be in terms of energy, computation cycles, monetary values or other resources~\cite{8952178}.

Tactic volatility frequently has a significant impact on the system's decision-making process, as it can directly impact if and when a tactic is selected and executed~\cite{moreno2017adaptation,Moreno:2017:CMP:3105503.3105511}. For example, tactic latency volatility may lead to scenarios where tactics are begun too early or too late~\cite{moreno2017adaptation,Moreno:2018:FED:3208359.3149180}. Tactic cost volatility may result in a system implementing a tactic that is more expensive compared with a tantamount, but less expensive alternative. 

Problematically, the vast majority of state-of-the-art decision-making processes do not account for tactic volatility and merely assume that tactics have static, unchanging attributes~\cite{moreno2017adaptation}. This inhibits the system's ability to account for real-world variability in its decision-making process and has been shown to adversely impact the system's effectiveness, resiliency and even security~\cite{ferdowsi2019cyber,smith2017cyber,8952178}. Accounting for tactic volatility is challenging due to the diverse and variable environments that many systems operate within.

We propose a \emph{Tactic Volatility Aware} process with \emph{evolved Recurrent Neural Networks} (eRNN) (TVA-E) to address the limitations of current systems in properly accounting for tactic volatility~\cite{8952178}. The system first predicts anticipated tactic attributes using \emph{evolved Recurrent Neural Networks} (eRNNs). These predicted tactic attributes are then incorporated into the system's existing decision-making process, providing it with more accurate information. \emph{Uncertainty reduction tactics} (URT)~\cite{10.1145/3194133.3194144} are also integrated into the system's adaptation control loop. Uncertainty reduction tactics are actions that reduce uncertainty, and improve the quality and quantity of knowledge used in the decision-making process. The novelty of this work, compared other studies~\cite{8952178,moreno2017adaptation}, is that it is the first known effort to: I) Evaluate the use of a neuroevolution (NE) algorithm for accounting for tactic volatility, and II) Evaluate the use of Uncertainty Reduction Tactics for enabling a system to better account for tactic volatility. 

This work advances the state-of-the-art enabling self-adaptive processes to better account for tactic volatility by implementing uncertainty reduction tactics and evaluate their ability to assist in accounting for tactic volatility. This work also broadens the field of machine learning applications to address tactic volatility analysis and mitigation. Moreover, using Neural Architecture Search (NAS) methods to optimize neural structures and parameters has been well applied to feed-forward ANNs for tasks involving static inputs, including convolutional variants~\cite{mikkulainen2017codeepneat,stanley2002evolving,stanley2009hypercube}, for the automated design of recurrent neural networks~\cite{rawal2016evolving,camero2018low,camero2019specialized}, for robotic controllers and game playing~\cite{risi2015neuroevolution,silva2013dynamics,galassi2016evolutionary,cardamone2009line,metzen2008towards,silva2015odneat}, and for evolving RNNs for time series data prediction has not been well studied~\cite{stanley2019designing}. This work advances these efforts by applying evolving eRNN as a prediction mechanism, and investigating its benefits in accounting for tactic volatility. 

Our evaluation addresses the following research questions: {\bfseries RQ1:} \emph{\RQA} We demonstrate effectiveness of eRNN for predicting tactic volatility. {\bfseries RQ2:} \emph{\RQB} eRNN effectively leads to more effective system actions when compared with existing techniques. {\bfseries RQ3:} \emph{\RQC} We demonstrate the foundational benefits of uncertainty reduction tactics in self-adaptive systems, specifically helping to make more accurate predictions regarding tactic latency and cost.

\section{Problem Definition} 
\label{sec: problemdefinition}




The attributes of a tactic are frequently primary inputs into the system's \emph{utility} calculation and overall decision-making process~\cite{moreno2017adaptation,Moreno:2018:FED:3208359.3149180}. The attributes of a tactic will frequently experience \emph{tactic volatility}, which is any rapid or unpredictable change that exists within the attributes of a tactic~\cite{8952178}. Due to their prevalence~\cite{8952178,moreno2017adaptation}, the two forms of volatility addressed in our work are \emph{latency} and \emph{cost}. 

\subsection{Tactic Latency Volatility}
Tactic latency is defined as the amount of time from when a tactic is initiated until its effect is produced~\cite{Moreno:2018:FED:3208359.3149180,moreno2017adaptation}. Examples of tactic latency include the amount of time required to fully activate a VM, or the necessary time to perform a system operation such as transferring a file over a network. Research is beginning to demonstrate the benefits of accounting for tactic latency in the self-adaptive decision-making process~\cite{camara2014stochastic,moreno2017adaptation,Moreno:2018:FED:3208359.3149180}. Unfortunately, many state-of-the-art adaptation processes still consider tactic latency to be a static value, one that does not change~\cite{8952178}. However, real-world systems will frequently encounter scenarios that experience tactic latency volatility. 
Accounting for tactic latency volatility is imperative for several reasons including:

\begin{enumerate}[leftmargin=.3cm]
    \item \descStep{Determine the most appropriate tactic(s)}{The anticipated latency of a tactic will likely be a significant determining factor for choosing the most appropriate tactic(s) that should be implemented by the system~\cite{moreno2017adaptation}.}

    \item \descStep{Augment a slower tactic with a faster one}{Sometimes a system will use a faster, but less optimal tactic to supplement a more optimal, but slower tactic~\cite{moreno2017adaptation}. This will enable the system to begin to realize at least some of the benefits from an adaptation decision earlier, while it is waiting for the slower tactic to be ready. When determining when and if to augment a slower tactic with a faster one, understanding tactic volatility is essential for knowing when and which tactic(s) to use.} 

    \item \descStep{Understanding when to start a tactic}{A robust proactive system will determine when a tactic is required, and optimally begin the execution of the tactic with a sufficient amount of time enabling it to be ready when needed. A system that is unable to account for tactic latency volatility will likely begin tactics too early (incurring additional costs) or too late (not available when needed)~\cite{8952178,moreno2017adaptation}.}

    \item \descStep{Incompatible tactics, and tactics that must be run in unison or succession}{Certain tactics may not be run whenever another incompatible tactic is being executed. 
    Contrarily, there may be cases when two tactics must be executed in unison. 
    Properly anticipating tactic latency is imperative for both of these scenarios. If multiple tactics are incompatible, understanding when a tactic is expected to begin and conclude is imperative for planning the execution of conflicting tactics. If multiple tactics must be executed in unison and at least one of these tactics contains latency, then planning must be done to anticipate when to begin the execution of the tactics so they are both properly executed together. Tactics may also need to be run in succession of one another. 
}

\end{enumerate}

\subsection{Tactic Cost Volatility}
The definition of tactic cost will widely vary and is likely to be domain-specific. Some examples of tactic cost include the energy, computations, or monetary value necessary to complete the execution of a tactic. Estimating tactic cost will likely be a primary concern for a self-adaptive system, especially if there are defined resource limitations or the system simply has the goal of achieving the highest reward at the lowest cost~\cite{moreno2017adaptation}. 

\begin{enumerate}[leftmargin=.5cm]

    \item \descStep{Cost may be a determining factor when selecting between multiple tactic options}{A tactic's cost may be the decisive factor for determining the most appropriate tactic that a system should select. Therefore, to determine the most appropriate tactic option(s), an accurate cost estimation is imperative.}

    \item \descStep{The anticipated cost may impact the system's ability to perform subsequent or concurrent tactics}{Systems frequently have a finite amount of resources. Therefore, it is imperative to accurately predict the cost of an action to: I) Determine if there are enough resources to perform the examined action, II) Select tactics to accrue the highest amount of possible utility.}

    \item \descStep{Cost may exceed reward}{Systems will frequently determine the expected cost of performing an operation and, in conjunction with the reward, determine if the operation is worth performing. If the cost exceeds the reward, then it may not be optimal for the system to execute the tactic.}


\end{enumerate}

\subsection{Uncertainty Reduction Tactics} 
\label{sec: UncertaintyReductionTactics}

Tactics are frequently used by self-adaptive systems to respond to events and accomplish system goals~\cite{moreno2017adaptation,4221625}. \emph {Uncertainty reduction tactics} (URT) are tactics that are specifically designed to reduce a system's uncertainty, and improve the quality and quantity of knowledge used in the decision-making process~\cite{10.1145/3194133.3194144}. Example uncertainty reduction tactics include probing a sensor that has not recently been used to determine its response time and to ensure that it is still active/available. Depending on the domain, numerous forms of data can be attained from uncertainty reduction tactics, some of which include information regarding the availability, response time and reliability of a resource. 
Uncertainty reduction tactics differ from conventional tactics in that the only objective of an uncertainty reduction tactic is to gather data and reduce uncertainty. An advantage of uncertainty reduction tactics is that they can provide additional meta data regarding an operation, typically without necessitating the cost/risk of executing the operation itself. 
While various uncertainty reduction tactics have been proposed, to our knowledge no existing works have implemented or evaluated uncertainty reduction tactics. We hypothesize that uncertainty reduction tactics can augment predictive processes such as eRNN to help to make more accurate tactic-based predictions and therefore positively impact the system's decision-making process. 


\subsection{Evolved Recurrent Neural Networks}
\label{sec:ernn}

Recurrent neural networks (RNNs) generally outperform classical statistical methods, \eg those from the auto-regressive integrated moving average (ARIMA) family of models, on data that is highly non-linear, acyclic and not seasonal. Further, classical statistical methods are not well suited to time series forecasting which incorporates multiple correlated time series of input data. This makes the use of RNNs a much stronger choice for providing predictions for self-adaptive systems. 

Hand-crafting effective and efficient structures for RNNs is a difficult, expensive, and time-consuming process. Instead of simply evaluating standard RNN architectures, neuro-evolution can be used to select, connect, and combine potential architectural components -- yielding a more rigorous and comprehensive search over available architectures, automating the design and training process. 

For this study, the Evolutionary eXploration of Augmenting Memory Models (EXAMM) algorithm~\cite{ororbia2019examm} was selected for the neuro-evolution process for several reasons. First, EXAMM progressively grows larger ANNs starting from a minimal seed architecture, in a manner similar to the popular Neuro-Evolution of Augmenting Toplogies (NEAT) algorithm~\cite{stanley2002evolving} that makes it tend to generate smaller and more efficient architectures. Furthermore, in contrast to NEAT, EXAMM utilizes higher-order node-level mutation operations, Lamarckian weight initialization (or the reuse of parental weights), and backpropagation through time (BPTT) to conduct local search, the combination of which has been shown to speed up both ANN training as well as the overall evolutionary process~\cite{desell2018accelerating}. EXAMM also operates with an easily-extensible suite of memory cells, not found in other strategies, including LSTM, GRU, MGU, UGRNN, $\Delta$-RNN cells and, more importantly, has the natural ability to evolve deep recurrent connections over large, variable time lags. Other research has also demonstrated that EXAMM has been shown to more quickly and reliably evolve RNNs in parallel than training traditional layered RNNs sequentially~\cite{desell-evostar-2019,10.1145/3377930.3390193,elsaid2020neuroevolutionary}.

\subsection{Motivating Example} 
\label{sec: MotivatingExample}


A cloud-based multi-tier web application with a self-adaptive component will serve as the motivating example. The system's goal is to maximize utility while minimizing cost, where the application is comprised of Virtual Machine (VM) servers that process incoming requests. Each VM instance has a defined cost and VMs can be added or removed from the resource pool as user demand dictates. Optional content (\eg advertisements can be reduced using the `dimmer' feature to efficiently provide content while encountering variable workloads. This example is based on work by Moreno~\cite{moreno2017adaptation}.

The target response time ($T$) and utility ($U$) is calculated (Equation~\ref{eq: motivatingExample}) as defined by the \emph{Service Level Agreement} (SLA). Penalties are incurred if the target response time is not met, while rewards are accrued for meeting the target average response time against the measurement interval. The average response rate is $a$, the average response time is $r$, the maximum request is $k$ and the length of each interval is defined as $\tau$. A dimmer ($d$) reduces provided content as necessary and cost ($C$) is proportional to the number of active VMs. In this example, the reward of the optional content ($R_O$) produces a higher utility than mandatory content ($R_M$). 

\begin{equation} \label{eq: motivatingExample}
	U = \left\{ \begin{array}{rl}
 	(\tau a (dR_O+(1-d)R_M)/C~~~~r\leq T \\ 
  	(\tau \min (0,a-k)R_O)/C~~~~r> T 
       \end{array} \right.
\end{equation}

Two tactics can be used to account for increases in user traffic: I) Add an additional VM server, or II) Use the \emph{dimmer} to reduce the proportion of responses that include the optional content. The tactic of adding a new VM server can take several minutes, while reducing optional content has negligible latency.

\smallTitle{Tactic latency volatility} If the system anticipates that the response time threshold will be surpassed in the immediate future, then it could proactively start the tactic of adding a VM server to keep the response time under the defined threshold. If the system overestimates latency, then it could incur superfluous additional costs since the VMs would be active longer than necessary. If the system determines that the defined response time threshold is going to be surpassed before a new VM server can be added, then an appropriate action may be to use the faster tactic of reducing optional content while the new VM server is being added. Inaccurate tactic latency predictions can result in scenarios where the system executes a tactic too early or too late, or even selects the improper tactic for the encountered scenario. Therefore, \emph{accounting for tactic latency volatility is a paramount concern}.

\smallTitle{Tactic cost volatility}Accounting for tactic cost volatility is important in this scenario since cost is an integral piece of the utility calculation. If cost is defined to be lower than what is being actually being regularly encountered, then this could result in scenarios where optional ($O$) content is shown too frequently. Conversely, if cost is defined to be higher than is being routinely encountered, then this could result in inaccurate utility calculations and scenarios where optional ($O$) content is shown too infrequently (leading to less reward). \emph{A volatility aware solution that enables the system to more accurately predict cost would enable the system to make decisions that lead to more optimal outcomes.}

\smallTitle{Uncertainty Reduction Tactics}
\emph {Uncertainty reduction tactics}~\cite{10.1145/3194133.3194144} are system actions that are specifically designed to reduce a system's uncertainty, and improve the quality and quantity of knowledge used in the decision making process. 
Example uncertainty reduction tactics include probing a sensor that has not recently been used or pinging a remote resource to ensure that it is still active. 
Uncertainty reduction tactics can provide valuable insight into predicting the latency and cost of executing a tactic such as activating an additional VM. While various uncertainty reduction tactics have been proposed, to our knowledge no existing works have implemented or evaluated uncertainty reduction tactics in the autonomous decision making process. \emph{Uncertainty reduction tactics may provide a lightweight, supportive input mechanism into making more accurate predictions regarding tactic volatility.}

\section{Proposed Tactic Volatility Aware Process}
\label{sec:proposedTechnique}

Most self-adaptive processes used for autonomous intelligent decision making employ a form of an \emph{adaptation control loop}~\cite{garlan2004rainbow}. An adaption loop cycles through all possible adaptation options at defined intervals, selecting the \emph{adaptation strategy(s)} that will return the highest utility. An adaptation strategy is a predefined composition of adaptation tactics (benefit)~\cite{moreno2017adaptation}. Using the motivating example in Section~\ref{sec: MotivatingExample}, during each adaptation control loop the system selects the adaptation strategy that is expected to return the highest utility. As described in Section~\ref{sec: problemdefinition}, predicting the volatility of the tactic is likely a crucial component of the decision-making process due to the impact that these predictions can have on the system's effectiveness, resiliency and ability to complete system and mission critical operations~\cite{8952178}. Our TVA-E process does not represent a new adaptation decision-making process, but provides more accurate information to existing adaptation processes.



\algnewcommand\algorithmicforeach{\textbf{for each}}
\algdef{S}[FOR]{ForEach}[1]{\algorithmicforeach\ #1\ \algorithmicdo}

\begin{algorithm} 
\caption{Integration of TVA-E into Adaptation Loop}
\label{alg:ProcessWorkFlow}

\small
\begin{flushleft}

    \hspace*{\algorithmicindent} \textbf{Input:} Time series data \\
    \hspace*{\algorithmicindent} \textbf{Output:} Adaptation decision

\end{flushleft}

\begin{algorithmic}[1]
\footnotesize
\item $  ${---Existing adaptation control loop---} \label{lst:line:ExistingAdapationControlLoop}
\Procedure{Adaptation Determination}{}
\label{lst:line:AdaptationDetermination}
    \ForEach {Adaptation Option} \label{lst:line:loopAllAdaptationOptions}
        \State $\hat{TacticVariables} \gets prediction(TimeSeriesData)$ \label{lst:line:TacticPrediction}  
        \State $\hat{predictedUtility} \gets UtilityCalc(\hat{TacticVariables})$ \label{lst:line:predictUtility}
    \EndFor

    \State $adaptationToExecute \gets getmax(predictedUtility)$ \label{lst:line:getMaxTacticExecute}
    
    \State $executeMaxAdaptation(adaptationToExecute);$ \label{lst:line:ExecuteMaxTactic}
    
    \State $TimeSeriesData \gets ObservedTacticAttributes()$ \label{lst:line:recordTacticObservation}
    \State $TimeSeriesData \gets UncertaintyReductionTactic()$ \label{lst:line:recordURTObservation}
\EndProcedure
\item $  ${---Existing adaptation control loop---}
\end{algorithmic}
\end{algorithm}


An overview of our TVA-E process is shown in Algorithm~\ref{alg:ProcessWorkFlow}. TVA-E integrates into the system's existing adaptation control loop (\texttt{L\ref{lst:line:AdaptationDetermination}}), iterating through all existing adaptation options (\texttt{L\ref{lst:line:loopAllAdaptationOptions}}). Predictions regarding the attributes of the tactic (\eg latency, cost, \etc) are performed using existing time-series data (\texttt{L\ref{lst:line:TacticPrediction}}). These predictions are then incorporated into the system's utility equation to determine the `goodness' of each adaptation option (\texttt{L\ref{lst:line:predictUtility}}). This anticipated utility then serves as a primary input to the system's decision-making process (\texttt{L\ref{lst:line:getMaxTacticExecute}}). For example, depending on the system it may select the strategy with the highest utility 
(\texttt{L\ref{lst:line:ExecuteMaxTactic}}). After the system executes the tactic(s), the observed tactic attributes are recorded as time-series data (\texttt{L\ref{lst:line:recordTacticObservation}}). Information recorded from the execution of the uncertainty reduction tactic is also stored as time-series data for future predictions (\texttt{L\ref{lst:line:recordURTObservation}}).

We chose eRNN for our TVA-E process due to its demonstrated ability to make accurate and reliable predictions using limited amounts of data in comparison with alternative options~\cite{desell-evostar-2019,10.1145/3377930.3390193,elsaid2020neuroevolutionary,elsaid2020neuro}. However, a myriad of alternative prediction techniques could be easily incorporated into our process to account for tactic volatility. 

Our proposed TVA-E process will easily integrate into popular adaptation processes such as Proactive Latency-Aware (PLA)~\cite{moreno2017adaptation} and MAPE-K~\cite{de2010software}, which is the most influential reference control model for autonomic and self-adaptive systems~\cite{elgendi2019protecting}. Integration into existing adaption processes is supported by using existing data to perform tactic predictions, and then providing the improved tactic predictions to the system's existing decision-making process. For example, the proposed TVA-E process will integrate into the $Analyze$ component of MAPE-K, providing enhanced tactic values to the $Plan$ component. This makes TVA-E highly applicable to a large number of existing self-adaptive processes and systems ranging from simple cyber-physical systems to large UAVs.

\section{Tactic Simulation Tool and Dataset}
\label{sec: ToolAndData}

An additional contribution of this work is the creation of the \emph{CELIA} (taCtic EvaLuator sImulAtor) tool. The objective of CELIA's simulation component is to regularly collect an updated version of a file from the remote location, with the goal of maximizing utility. CELIA uses the utility equation: $\hat{U}=(\frac{Reward}{(\hat{Latency}+\hat{Cost})})$ to determine if the expected reward in relation to the expected latency and cost of the operation warrants the file collection and extraction operations, or if the system should `pass' and wait for another decision-making iteration. The expected reward is compared against a pre-determined threshold value according to a system's Service Level Agreement (SLA). When determining if the system should perform tactic operations, the simulation component takes two files as input: I) The predicted values for each tactic attribute (latency, cost), and II) The observed tactic attributes which serve as the ground truth. CELIA iterates through each adaptation decision-making cycle first using the predicted tactic attributes, and then the actual values (the ground truth). CELIA provides several output values, such as the absolute difference of the observed (ground truth) vs. expected utility and if the adaptation decision was correct. 

CELIA emulates a simple intelligent self-adaptive system that is tasked with downloading and processing a file from multiple remote locations. It repeatedly performs the following: I) download a compressed file from remote servers, emulating the tactic of performing a communication or file transmission operation; II) extract the file's contents, emulating the tactic of performing a simple file I/O operation; III) perform a {\tt grep} of the extracted contents; IV) compress the extracted files contents, providing an additional tactic example of a file I/O operation; V) delete the file thus providing an additional example of a file i/o tactic; and VI) A simple uncertainty reduction tactic that gathers augmenting information regarding the availability and response time of the remote resource. The data created by CELIA is important since existing resources such as `The Internet Traffic Archive'\cite{TheInternetTrafficArchive_URL} do not contain both latency and cost values for performed actions. Our created data, source code and Docker image is provided on the project website~\cite{CELIA_URL}.

For this work, the `download' tactic perpetually downloads the Apache installation file\footnote{\url{https://downloads.apache.org/httpd/httpd-2.4.43.tar.gz}} from three real-world hosting servers located around the world (USA, Switzerland, Canada). Servers hosting Apache were chosen due to their prevalence and that they were anticipated to experience real-world volatility. Variability and uncertainty are inherently included in these operations since they are being conducted using `live' servers located across the world, providing impacts such as periodic latency and communication challenges just as encountered in any real-world operation. Tactic latency is the amount of time necessary to perform each operation, while tactic cost is average CPU resources used for the operation. Overall, \TotalTacticOneServerOneEvals tactic operations were used in our evaluation. 

We chose to evaluate the ability of the examined methods using this tactic as it contained the most real-world volatility, emulating the actions that a file transfer operation would resemble in a self-adaptive system. This tactic consistently experienced real-world volatility due to numerous factors such as network and server variability. Additional reasons for focusing on this tactic were that: I) The latency and cost variability provided by this server was sufficient to provide a proper evaluation, and II) To simplify the analysis output/examination.

\section{Experimental Design}
\label{sec:experimental_design}

To determine the effectiveness of eRNN in relation to alternative options, we evaluated the following prediction mechanisms against our eRNN solution: AutoRegressive Integrated Moving Average (ARIMA), Support Vector Regressor with a Radial Basis Function kernel (SVR-RBF), one layer Multi-layer Perception (MLP, \eg feed forward neural networks) and Long Short-term Memory (LSTM) recurrent neural networks. We compared our eRNN-based technique against these techniques~\cite{ahmed2010empirical} for a variety of reasons. ARIMA was used in a similar work~\cite{8952178}, while LSTM RNNs are well known architecture for forecasting multivariate time-series data because of their gated memory cells which utilize information from previous time steps~\cite{petnehzi2019recurrent}. MLP are basic feed forward neural networks that consist of only feed forward fully connected hidden layers, therefore serving as a baseline neural network model. We also included SVR as it is a sophisticated model which has been shown to perform well on time series data prediction~\cite{lin2007time,muller1997predicting}. Each model was provided a total of \TotalTacticOneServerOneEvals records (36,472 records for training, 7,819 for testing and 7,815 for validation).

For our experiments, we used ARIMA (1, 1, 0), with a differencing order of d = 1 as the time-series data was non-stationary. Autocorrelation plots were used to determine autoregressive (p=1) and moving terms (q=0). Support vector regression (SVR) was performed with a radial basis function (RBF) kernel and a linear kernel. The regularization parameters and hyper-parameters of the kernel function were set to the default values of 1.0 and 0.1. The MLP model was fully connected with one hidden layer of 100 nodes with the output layer predicting cost and latency. The LSTM model was fully connected, consisting of a single hidden layer with 1,000 LSTM nodes followed by the output layer. 

The neural networks and EXAMM processes trained the neural networks using the training data and utilized the testing data to optimize hyperparameters and determine when to stop the training process. The resulting eRNNs were quite small, with the largest network having only 44 total nodes and 590 weights. The validation data was excluded from the evolutionary process and only used at the conclusion to determine the final performance and make sure the trained networks were generalizable.

After demonstrating the benefits of eRNN compared to other evaluated machine learning options, we then assessed the benefits of uncertainty reduction tactics used in conjunction with our eRNN-based process. We evaluated URTs with eRNN only since eRNN was found to be the most effective prediction mechanism that was able to utilize URTs. ARIMA is a univariate model and is unable to leverage information from other sensors (URT \# 1). This evaluation was accomplished by implementing two uncertainty reduction tactics that were chosen due to their real-world applicability~\cite{van2014opennetmon} and discussion in a previous work~\cite{10.1145/3194133.3194144}. The two URTs used in our evaluation were:

\newlist{learningObjectives}{enumerate}{1}

\setlist[learningObjectives,1]{label={},noitemsep, leftmargin=.13in}%
\newcommand{\LearningObjective}[3]{#1: #2 (#3)}

\newlist{URTItems}{enumerate}{1}
\setlist[URTItems,1]{label={},noitemsep, leftmargin=.05cm}%
\newcommand{\URTItem}[3]{\textbf{#1: #2 -} #3}


\begin{URTItems}
    \item \URTItem{URT \#1}{Reducing uncertainty due to model drift}{The models used may progressively become inconsistent with the system's environment due to various forms of internal and external volatility. This uncertainty can be addressed by incorporating additional sensors or data gathering operations~\cite{10.1145/3194133.3194144}. In our evaluation, we emulate this uncertainty reduction operation by regularly `pinging' the remote server to attain information about its availability and response time. This information is then incorporated into the system's tactic prediction process.}\vspace{2mm}

    \item \URTItem{URT \#2}{Changing the sampling rate of a parameter}The mean and variance of the observations can generate a confidence interval for the monitored parameter value. Adapting the sampling rate is one method of controlling the width, and uncertainty of the interval~\cite{van2014opennetmon}. We emulate such an uncertainty reduction operation by incorporating every \textit{n}th observation into our prediction process. The \textit{n}th value would be determined according to system specifications. For this experiment, we chose \textit{n} as 5, 10, 20. These values were chosen to mimic loss of information that would occur by changing sampling rate with respect to the size of our dataset. Values less than 5 did not represent a significant reduction in sample size due to the magnitude of our data. We also chose a sampling rate of 10 and 20 by doubling the previous sampling rate. Our data didn't allow for values beyond 20 because the amount of data was not sufficient for our experiments.
    
    URT \#2 is a flexible framework that allows us to adjust the proper URT strategy to specific learning tasks. In this experiment, we conduct extensive cross-validation and choose n=5 for latency prediction and n=20 for cost prediction for optimal prediction improvement. The advantage of such flexibility may not be significant in this study since the utility is only determined by two factors (latency and cost). However, if we consider a more complex simulation environment where the utility depends on hundreds or thousands of factors, we can use customized sampling rates to minimize the uncertainty in each factor input resulting in a considerable impact on the utility score.

\end{URTItems}

The following evaluation criteria was used in our analysis:

\begin{enumerate}[leftmargin=.35cm]
 
     \item \descStep{Ability to accurately predict tactic attributes}{The predicted attributes of a tactic frequently have a significant impact on the system's decision-making process, as it can directly impact if and when a tactic is selected and executed~\cite{moreno2017adaptation,Moreno:2017:CMP:3105503.3105511}. Therefore, it is important for a system to accurately predict the attributes of a tactic to ensure effective, efficient and resilient functionality~\cite{8952178}. In our evaluation, we compared the predicted tactic latency and cost values against the observed ground truth values.}
 
    \item \descStep{Expected vs achieved utility}{We evaluated the system's ability to achieve the predicted amount of utility while operating in dynamic environments with variable data. Accurately anticipating the achieved utility is imperative for a system's decision-making process for choosing the most appropriate tactic(s) and adaptation strategies for the encountered scenario~\cite{Moreno:2017:CMP:3105503.3105511,Moreno:2018:FED:3208359.3149180,CMU-ISR-17-119,moreno2017adaptation}. In our evaluation, we compared the predicted utility vs. the observed utility (ground truth).}

    \item \descStep{Ability to make correct decisions}{The ability to make proper decisions is paramount for self-adaptive systems, therefore we recorded when the system: {\it a)} Performed an action when the system \ul{should not} have, {\it b)} Performed an action when the system \ul{should} have, {\it c)} Didn't perform an action when it \ul{should not} have, and {\it d)} Didn't perform an action when it \ul{should} have.
    

    Our evaluation compared the predicted recommended system action vs the action that the system should have taken using observed values (ground truth). We evaluated the impact of each prediction mechanism on a system's decision-making process both quantitatively and qualitatively.}

\end{enumerate}





\pgfplotstableread{

}\ernnWlatencytable

\section{Experimental Results}
\label{sec:experimental_results}


\noindent \textbf{RQ1: \RQA} This research demonstrates the capabilities of eRNN in both the general prediction process, and more specifically in benefiting systems in accounting for tactic volatility. For tactic latency and cost, we first use the mean squared error (MSE) to evaluate the predicted result, as it is a common loss function for regression tasks. However, MSE is known to be sensitive to the scale of the data and the outliers. Therefore, we also reported the Mean absolute error (MAE) as an alternative error measurement.
The two measurements are given by: $MSE=\frac{1}{N}\sum_{i=1}^{N}(t_i-\hat{t}_i)^2$, and $MAE=\frac{1}{N}\sum_{i=1}^{N}|t_i-\hat{t}_i|$.
, where $N$ is the total number of test cases, $t$ is the ground truth value (\eg latency or cost) and $\hat{t}_i$ is the predicted value. ARIMA utilizes univariate data to make predictions, therefore using data from uncertainty reduction tactics in addition to cost or latency data is not possible (and is therefore not reported). We ran each model on the three different download servers representing various tactic options (Section~\ref{sec: ToolAndData}). We report the averaged performance measurements of the three servers ($\overline{MSE}$, $\overline{MAE}$) in Table~\ref{table:evaluationPredictedLatency}.

\begin{table}[t]
\small 
\begin{center}
\caption{eRNN's capabilities for predicting tactic latency.}
\label{table:evaluationPredictedLatency}
\begin{tabular}{l|l|l|l}
\multirow{2}{*}{\bfseries URT tactic} & \multirow{2}{*}{\bfseries Model} &
\multirow{2}{*}{$\bm{\overline{MSE} \times 10^{-3}}$} & \multirow{2}{*}{\bfseries $\bm{\overline{MAE} \times 10^{-1}}$} \\
 &  &  &  \\ \hline
\multirow{6}{*}{no URT} & \textbf{eRNN} & \textbf{0.61} & \textbf{0.16} \\ \cline{2-4} 
 & ARIMA & 0.69 & 0.17 \\ \cline{2-4} 
 & LSTM & 0.83 & 0.16 \\ \cline{2-4} 
 & MLP & 0.86 & 0.16 \\ \cline{2-4} 
 & SVR\_linear & 8.87 & 0.35 \\ \cline{2-4} 
 & SVR\_rbf & 9.0 & 0.36 \\ \hline
\begin{tabular}[c]{@{}l@{}}URT\_Combine\\ (Sample rate=5)\end{tabular} & \textbf{eRNN} & \textbf{0.57(7\%$\uparrow$)} & \textbf{1.15(6\%$\uparrow$)} \\ 
\bottomrule
\end{tabular}
  \end{center}
\end{table}

\begin{figure*}[!htb]
\centering
\begin{minipage}{.45\textwidth}
\centering
\begin{tikzpicture}[scale=.7]
\begin{groupplot}[
boxplot/draw direction=y,
ylabel={Cost},
xtick={1,2,3},
width=\textwidth,
height=7.45cm,
ymin=0, ymax=0.3, 
boxplot/box extend=0.1,
xticklabels={Ground Truth, eRNN w/o Uncert Red, eRNN w Uncert Red},
    x tick label style={
        text width=1.5cm,
        align=center,
    },
    y tick label style={
      /pgf/number format/precision=2,
      /pgf/number format/fixed},
      group style={
        group size=3 by 3,
          horizontal sep=0cm,vertical sep=0cm
      },
]

\nextgroupplot
\addplot+[boxplot,mark size=1pt]
table[row sep=\\,y index=0] {\groundtruthcosttable};

\addplot+[boxplot,mark size=1pt]
table[row sep=\\,y index=0] {\ernnWOcosttable};
\addplot+[boxplot,mark size=1pt]
table[row sep=\\,y index=0] {\ernnWcosttable};

\end{groupplot}
\end{tikzpicture}
\caption{Predicted tactic cost (on server1)}
\label{fig:Cost}
\end{minipage}
\begin{minipage}{.45\textwidth}
\centering
\begin{tikzpicture}[scale=.7]
\begin{groupplot}[
boxplot/draw direction=y,
ylabel={Latency},
xtick={1,2,3},
width=\textwidth,
height=7.45cm,
ymin=0, ymax=0.15, 
boxplot/box extend=0.1,
xticklabels={Ground Truth, eRNN w/o Uncert Red, eRNN w Uncert Red},
    x tick label style={
        text width=1.5cm,
        align=center,
    },
    y tick label style={
      /pgf/number format/precision=2,
      /pgf/number format/fixed},
      group style={
        group size=3 by 3,
          horizontal sep=0cm,vertical sep=0cm
      },
]
\nextgroupplot
\addplot+[boxplot,mark size=1pt]
table[row sep=\\,y index=0] {\groundtruthlatencytable};
\addplot+[boxplot,mark size=1pt]
table[row sep=\\,y index=0] {\ernnWOlatencytable};
\addplot+[boxplot,mark size=1pt]
table[row sep=\\,y index=0] {\ernnWlatencytable};
\end{groupplot}
\end{tikzpicture}
\caption{Predicted tactic latency (on server1)}
\label{fig:Latency}
\end{minipage}
\end{figure*}

\begin{itemize}[leftmargin=.3cm]

\item Table~\ref{table:evaluationPredictedLatency} 
demonstrate that eRNN achieves the best performance for predicting latency and cost. The uncertainty reduction tactic (URT) doesn't help eRNN to improve the predictive power in terms of MSE, but does help with MAE.

    
    \item Figure~\ref{fig:Latency} demonstrates that the predicted latency has a large number of outliers in the observed values (ground truth). To better evaluate the model in such a case, we further report the MAE for each compared model as it is less sensitive to outliers. A small percentage of outliers will likely have little impact on the functionality of the system from a more aggregate level (\eg it doesn't matter if the prediction that leads to an non-optimal action is only slightly wrong or very wrong, the incorrect action was still taken). In RQ \#2, we further justify our claim that the higher MAE does not prevent eRNN from making effective decisions. 

    \item The additional uncertainty reduction tactic information has a significant impact on eRNN for cost prediction. The MSE is reduced by approximately half when uncertainty reduction tactic information trains the eRNN.

    \item The additional uncertainty reduction tactic information for model training serves as regularization. Figure~\ref{fig:Cost} and Figure~\ref{fig:Latency} demonstrate that without uncertainty reduction tactic information, eRNN tends to provide more predictions that are close to the outliers (which are likely to be nothing more but systematic noises). Such an overfitting issue is well addressed with the presence of uncertainty reduction tactic information.

    \item eRNN provides significantly more efficient networks. The MLP models had 900 and 1,100 trainable parameters (without and with URT \#1), the LSTM models had 19,000 and 21,000 trainable parameters (without and with URT \#1), and the eRNN models had 485 and 590 weights (without and with URT \#1). EXAMM was able to find RNNs that not only had better performance, but were orders of magnitude smaller in terms of trainable connections -- highlighting the inefficiency of using traditional fixed architectures. These evolved networks have the potential to more easily be embedded in systems with computational and power limitations, which is common in adaptive tactical scenarios~\cite{knudson2011adaptive,bailey2014adaptive}.
\end{itemize}

\noindent \textbf{Outcome:} \textit{Our findings demonstrate that eRNN is effective at both making predictions, and specifically at predicting tactic cost and latency volatility.}



\noindent \textbf{RQ2: \RQB} Unlike the tactic cost and latency whose values are generally distributed around 1, the calculated utility’s value is not well scaled (range from 0 to 16,441.9). As a result, we use the scale-independent measurement, mean absolute percentage error (MAPE), instead of MSE to evaluate the utility prediction. The formula for MAPE is given by: $MAPE=\frac{1}{N}\sum_{i=1}^{N}|\frac{t_i-\hat{t}_i}{t_i}|$.

In Table~\ref{table:evaluationofUtility} the MAPE exceeding 100\% indicates that the model is likely to heavily overestimate the utility especially when the true utility is small (close to zero). This is usually considered a significant prediction failure. In self-adaptive systems, the large prediction failure has a high risk of leading the incorrect adaptation decisions. We found that eRNN performs similarly to state-of-art methods such as ARIMA, LSTM and MLP, and that its performance can be further boosted with URT operations.

Accurately anticipating utility is imperative for systems to choose the most appropriate tactic(s) and adaptation strategies for the encountered scenario~\cite{Moreno:2017:CMP:3105503.3105511,Moreno:2018:FED:3208359.3149180,CMU-ISR-17-119,moreno2017adaptation}. The benefits can be further improved by leveraging the uncertainty reduction data during the training phase, specifically for eRNN for which the URT produces a 24\% improvement. 

\begin{table}[]
\small 
\begin{center}
\caption{Evaluation of utility prediction\ demonstrating the how predicted utility deviates from the true utility.}
\label{table:evaluationofUtility}
\begin{tabular}{l|l}
\bfseries Model & $\bm{\overline{MAPE}}$ \\ \hline
eRNN & 21\% \\ \hline
ARIMA & 13\%\\ \hline
LSTM & 16\% \\ \hline
MLP & 14\% \\ \hline
SVR\_linear & 57\% \\ \hline
SVR\_rbf & 57\% \\ \hline
\begin{tabular}[c]{@{}l@{}}eRNN+URT\_Combine\\ (Sample rate=5)\end{tabular} & 16\% \\ 
\bottomrule
\end{tabular}
  \end{center}
\end{table}

The final decision-making process can be seen as a binary classification task. Thus, we can evaluate the system as a binary classifier. We begin by defining several terms. The true positive (TP) represents the number of correct `update' decisions made by the system. The true negative (TN) is the correct `not update' decisions made by the system. The false positive (FP) is the number of incorrect `update' decisions and the false negative is the number of incorrect `not update' decisions. We can then evaluate the classification accuracy of the system output. This accuracy calculation reveals the proportion of the correct decisions in the system output. For the wrong decisions, we use false positive rate (FPR) and false negative rate (FNR) for evaluation where $FPR=\frac{FP}{TP+FP}$ is the frequency of wrong `update' and $FNR=\frac{FN}{FN+TN}$ is the frequency of wrong `not update'. Using the values from Table~\ref{table:evaluationofDecisionMaking}, we can conclude:

\begin{itemize}[leftmargin=.3cm]
    \item The uncertainty reduction information helps improve the accuracy by 13\% in our eRNN-based TVA-E) technique, and achieves the highest accuracy.

    \item The eRNN-based process has the lowest FPR. This is consistent with the previous observation on its small MAE of the utility prediction. Since our eRNN-based system rarely overestimates the small utility values, it is less likely to produce a false update tactic action.

    \item The eRNN-based system has moderate FPR and FNR, but results in low prediction accuracy compared with other baselines. We observe that URTs manage to rebalance the two types of mistakes (\ie FPR and FNR) so that the overall benefit (\ie the accuracy) can be maximized. In the rebalanced decision making procedure, URT chooses to tolerate the increase of FNR for this type of mistake will only have a minor negative impact if the amount of the utility missed by incorrect `not update' decisions are small. We can use utility loss to quantify such negative impact. The utility loss is the sum of the ground truth utility scores for all incorrect `not update' decisions. Intuitively, the utility loss measures the amount of utility a system could have gained with the correct `update' decisions. eRNN also significantly reduced the utility loss compared to other strategies, obtaining results two orders of magnitude better than the next best strategy (MLP).

    \item The high FNR rate of eRNN based system has a minor negative impact on the final decision-making. The false-negative decisions made by eRNN based system have the least utility loss compared with other methods. 

    \item Our eRNN-based TVA-E process can leverage uncertainty reduction information to reduce the utility loss to 0.1\%, meaning that UCT helps eRNN to minimize the expected loss during the decision making process. Similar to utility loss, we can use utility gain to quantify the positive impacts made by correct decisions. The utility gain is the sum of the true utility scores for all correct `update' decisions.

\end{itemize}

\begin{table}[h!]
\small 

\begin{center}

\caption{Comparison of the decisions made by the simulated self-adaptive systems using evaluated prediction mechanisms (threshold=1000).}

\label{table:evaluationofDecisionMaking}
\begin{tabular}{l|l|l|l|l|l}
\bfseries Model & $\bm{\overline{FPR}}$ & $\bm{\overline{FNR}}$ & $\bm{\overline{U\_gain}}$ & $\bm{\overline{U\_loss}}$ & $\bm{\overline{Acc}}$ \\ \hline
\bfseries eRNN & 15\% & 7\% & 2.10E+07 & 1.40E+05 & 88\% \\ \hline
\bfseries ARIMA & 6.6\% & 5.6\% & 2.1E+07 & 9.3E04 & 94\% \\ \hline
\bfseries LSTM & 7\% & 11\% & 2.00E+07 & 1.90E+05 & 91\% \\ \hline
\bfseries MLP & 9.20\% & 6\% & 2.00E+07 & 9.00E+04 & 92\% \\ \hline
\bfseries SVR\_linear & 0\% & 54\% & 2.00E+07 & 1.80E+06 & 59\% \\ \hline
\bfseries SVR\_rbf & 0\% & 56\% & 2.00E+07 & 1.90E+06 & 58\% \\ \hline
\begin{tabular}[c]{@{}l@{}}\bfseries eRNN + \\\bfseries URT\_Combine\\ \bfseries (Sample rate=5)\end{tabular} & 0.53\% & 17\% & 2.30E+06 & 3.40E+02 & 99\% \\ 
\bottomrule
\end{tabular}
  \end{center}
\end{table}

Our general observations are that: {\it a)} eRNN is the best candidate process for latency and cost predictions, as the model performance is both accurate and robust. {\it b)} The strength of eRNN cost and latency prediction can be well transformed to utility prediction. {\it c)} Our eRNN-based process makes the most correct decisions and also maximizes the utility gain and minimize the utility loss during the decision-making. {\it d)} The uncertainty reduction information plays the role of regularizing the model prediction/risk control/reducing the uncertainty of the model output. {\it e)} Uncertainty reduction information can reduce utility loss.


\noindent \textbf{Outcome:} \textit{Our analysis demonstrates that eRNN is the best method for predicting volatility compared with existing prediction techniques. We also observe that uncertainty reduction tactics provide useful information to help make more accurate decisions.}


\noindent \textbf{RQ3: \RQC} While the potential benefits of uncertainty reduction tactics have been previously discussed~\cite{10.1145/3194133.3194144}, there have been no known efforts to demonstrate or evaluate their potential benefits in simulated self-adaptive systems. As demonstrated in RQ1 \& RQ2, we found that uncertainty reduction tactics were beneficial in helping eRNN to provide better tactic latency and cost predictions.  

\begin{itemize}[leftmargin=.3cm]

    \item The effectiveness of URT is model specific. While all the models exhibit improvements with URT added, eRNN best leverages the uncertainty information and improve its MSE by 50\% (on average, other methods with UTC provide less than 10\% improvement). 

    \item Additional uncertainty reduction tactic information for model training serves as regularization. Figure~\ref{fig:Cost} and Figure~\ref{fig:Latency} demonstrate that without uncertainty reduction tactic information, eRNN tends to provide more predictions that are close to the outliers. Overfitting issues are well addressed with the use of uncertainty reduction tactic information. 
\end{itemize}

\noindent \textbf{Outcome:} \textit{This work demonstrates the potential benefits of uncertainty reduction tactics, specifically in helping to make more accurate predictions regarding tactic volatility.}

\section{Discussion} 
\label{sec:discussion} 

Self-adaptive systems will be increasingly expected to efficiently, effectively, and resiliently perform in volatile environments. This work further demonstrates the need for self-adaptive systems to account for the volatility of tactics (tactic volatility). Demonstrated benefits of accounting for tactic volatility include reduced uncertainty and an increased probability of the system making optimal decisions. 

This work demonstrates the benefits of eRNN in predicting tactic volatility, and also the general capabilities of our eRNN-based TVA-E process as well. These demonstrated benefits exhibit the capabilities of eRNN and provides the foundation for its application into other areas of machine learning. Thus, the findings of this work advance both research in self-adaptive systems and in a multitude of other areas of applied machine learning.

While the benefits of uncertainty reduction tactics have been discussed from a theoretical perspective~\cite{10.1145/3194133.3194144}, this is the first effort to evaluate their benefit in a simulated system. This creates the foundation for future work, such as I) The examination of uncertainty reduction tactics in additional simulated and physical environments and settings, II) Their further evaluation, III) Creation of additional uncertainty reduction tactics in a multitude of areas and purposes. The demonstrated benefits of the supplementary information provided to eRNN by uncertainty reduction tactics demonstrates the benefits of this additional information to this model. Uncertainty is widely recognized as being detrimental~\cite{10.1145/3194133.3194144,camara2017reasoning}, and this work demonstrates the ability of uncertainty reduction tactics to diminish the amount of uncertainty encountered by the system.


The provided dataset and simulation tool~\cite{CELIA_URL} will assist other researchers and practitioners in creating and evaluating their own tactic volatility aware processes. The necessity of such processes will continue to increase as the recognized need for systems to effectively function in volatile environments correspondingly grows. 



\section{Conclusion and Future Work}
\label{sec: conclusion}

Like most uncertainty reduction tactics, this work did not account for cost or risk associated with real-world uncertainty, since these aspects are negligible by design. We recognize that the specific uncertainty reduction operations used within a system will be entirely domain specific and therefore it is impossible for any work to assume that it could implement and evaluate every form of uncertainty reduction operation. 
Future work will further investigate uncertainty reduction tactics, and EXAMM shall be used for continuous online evolution as new information is received to offer online accurate predictions.

We propose a new \emph{Tactic Volatility Aware} process with \emph{evolved Recurrent Neural Networks} (TVA-E). TVA-E is able to easily integrate into popular adaptation processes, enabling it to readily and positively impact a large number of self-adaptive systems. Our simulations using \TotalTacticOneServerOneEvals records demonstrate that: I) TVA-E can effectively account for tactic volatility, II) eRNN demonstrates its effectiveness compared with other leading machine learning alternatives, III) Uncertainty reduction tactics can be beneficial in accounting for tactic volatility. Complete results, evaluation software and other information is publicly available on the project website: \ToolURL

\bibliographystyle{ACM-Reference-Format}
\bibliography{tactics}

\appendix
\balance
\end{document}